\documentclass{isprs}
\usepackage{times}
\usepackage{epsfig}
\usepackage{graphics}
\usepackage{multirow}
\usepackage[cmex10]{amsmath} % define this before the line numbering.

\usepackage{pgfplots}
\usepackage{url}
\usepackage{setspace}
\usepackage{geometry} % added 27-02-2014 Markus Englich
\usepackage[list=true]{subcaption}

\usepackage[labelsep=period]{caption}

\newcommand{\SeeFig}[1]{Figure \ref{#1}}
\newcommand{\SeeSec}[1]{Section \ref{#1}}

\newcommand{\SeeEq}[1]{Equation \ref{#1}}
\newcommand{\SeeTable}[1]{Table \ref{#1}}

\begin{document}

\title{MultiCol-SLAM - A Modular Real-Time Multi-Camera SLAM System}

\author{S. Urban, S.Hinz}

\address
{
	Institute of Photogrammetry and Remote Sensing, Karlsruhe Institute of Technology  Karlsruhe \\
	Englerstr. 7, 76131 Karlsruhe, Germany - (steffen.urban, stefan.hinz)@kit.edu \\ http://www.ipf.kit.edu
}

\abstract
{
The basis for most vision based applications like robotics, self-driving cars and potentially augmented and virtual reality is a robust, continuous estimation of the position and orientation of a camera system w.r.t the observed environment (scene).
In recent years many vision based systems that perform simultaneous localization and mapping (SLAM) have been presented and released as open source.
In this paper, we extend and improve upon a state-of-the-art SLAM to make it applicable to arbitrary, rigidly coupled multi-camera systems (MCS) using the MultiCol model.
In addition, we include a performance evaluation on accurate ground truth and compare the robustness of the proposed method to a single camera version of the SLAM system.
An open source implementation of the proposed multi-fisheye camera SLAM system can be found on-line \url{https://github.com/urbste/MultiCol-SLAM}.

}


\maketitle

\section{Introduction}
\label{ch:tracking}
The accurate reconstruction of an observed scene from sets of ordered images has a long history in aerial \cite{kraus2004photogrammetrie} and close-range photogrammetry \cite{luhmann2006close}.
Usually, the object and reconstruction setup is well defined and the scene observations using high-resolution cameras are well planed.
Thus, the connectivity between multiple camera positions is known or easily established and off-line bundle adjustment over the cameras and scene structure is performed yielding accurate results.
In addition, initial values for the exterior and interior camera orientations are mostly available from external sensors, passpoints and accurate calibration.

In the computer vision community, the direction of research is called SfM and relaxes many constrains about scene and camera geometry that are assumed in classical photogrammetry.
Advances in projective geometry \cite{Hartley_and_Zisserman_2008} and visual feature research \cite{weinmann2013visual} enabled the off-line reconstruction of large scenes from unordered sets of images and photo collections \cite{wu2013towards,snavely2006photo,agarwal2009building,wu2011multicore,szeliski2010computer,sweeney2015optimizing,triggs2000bundle}.
But essentially, the SfM methods solve the exact same problem as in aerial and (close-range) photogrammetry, i.e. reconstruction of scene and cameras.
The main difference lies in the initialization of the bundle adjustment through direct relative orientation methods for calibrated \cite{stewenius2006recent,Hartley_and_Zisserman_2008,kneip2012finding} or uncalibrated \cite{barreto2005fundamental,kukelova2015radial} cameras and a simultaneous connectivity estimation using only natural image features.

In both communities, the scene is basically assumed to be static and the reconstruction is done off-line doing batch optimization over the entire scene.
Hence, correspondence information (features) can be exhaustively extracted and matched a-priori and has no temporal coherence.
Latter however, is the case for live video frames from a moving camera.
In addition, the pose change (baseline and orientation) between subsequent frames can be very small.

Thus, the robotics community, where sensors (including cameras) are mounted on top of moving platforms, developed filter-based methods \cite{davison2004real,montemerlo2002fastslam,davison2007monoslam,montemerlo2007fastslam} to estimate the sensor pose from continuous sensor updates (live video).
In recent years, basically two streams emerged from this line of research, namely filter- and keyframe-based SLAM techniques that will be described in more detail in the following.
The basic idea behind both approaches is that not all poses, observations and uncertainties from a continuous video stream can (and have to) be integrated into the solution of the SLAM problem.
Outstanding work of \cite{dellaert2006square} and \cite{strasdat2012visual} give a detailed analysis of both methods and close the gap between SLAM, SfM and classical bundle adjustment by generalization of the entire reconstruction problem using graphical models.

\begin{figure*}
	\centering
	%trim=l b r t
	\includegraphics[trim = 5mm 110mm 55mm 5mm, clip=true, width=\textwidth]{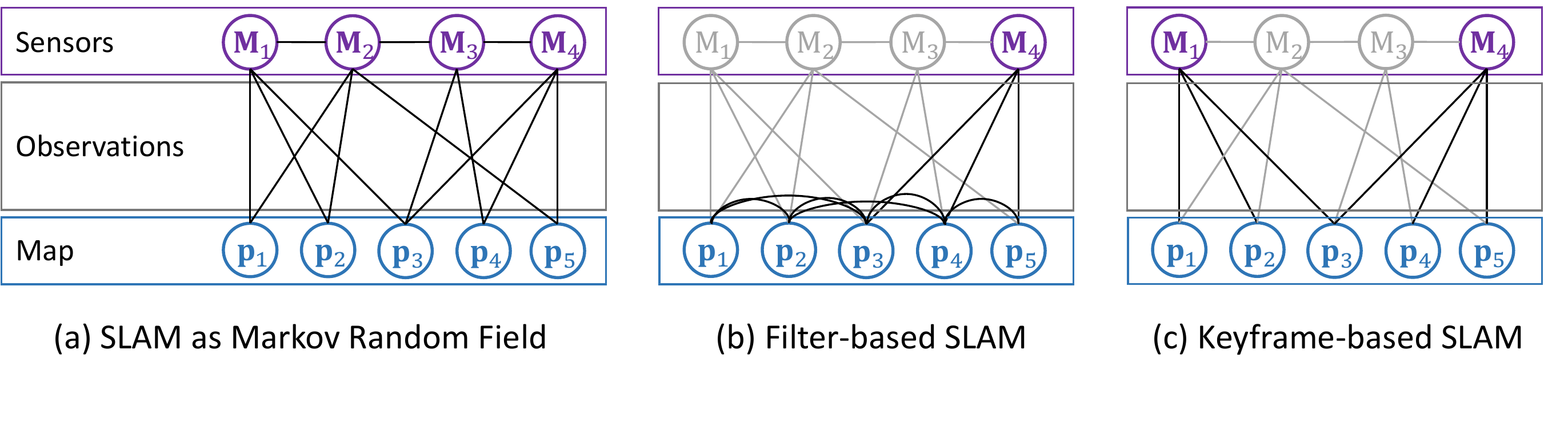}
	\caption{(a) SLAM as a Markov random field. (b) The graph is sparsified by marginalization of past poses except the current one from the graph. (c) Only keyframes are selected and all other poses, observations and object points, that are not visible in any retained keyframes are removed.}
	\label{fig:SlamProblem}
\end{figure*}

\SeeFig{fig:SlamProblem} depicts the structure in an undirected graph (Markov Random Field) for a toy example.
In total, four frames were recorded that observe a scene consisting of five map points.
The poses are connected to map points by edges (observations) and the poses themselves are connected by state transitions (e.g. inertial sensor).
An optimal BA solution is given (learned) by estimating the ML solution for the depicted graph and could be computed efficiently in this example.
Speaking in terms of graphical models, the ML solution corresponds to the joined probability distribution over all parameters (poses and map points).
Now, lets grow he graph by adding poses and map points.
With each frame, the computational complexity increases and quickly exceeds the runtime constrains for real-time applications.
Thus, we need a way of thinning out the graph.
Using a filter-based approach, historic poses are marginalized out of the joint distribution by integration over all other parameters.
The resulting graph is depicted in \SeeFig{fig:SlamProblem}b.
The issue with filter-based SLAM becomes visible.
To marginalize the current pose from the joint distribution, all potentials between connected variables need to be stored and updated with every frame and thus the number of map points will be very limited.

But especially a large map and many observed features lead to accurate and robust SLAM pipelines \cite{strasdat2012visual}.
Thus, the idea behind keyframe-based SLAM is to sparsify the graph by simply removing poses, map points and observations from the graph instead of marginalizing them from the probability distribution.
In this way, classical BA can be performed which is efficient and fast to compute on sparse matrices \cite{triggs2000bundle,kummerle2011g}

Thus in this paper, a keyframe-based SLAM pipeline will be used as the basis for the continuous estimation of map and MCS pose.
An additional benefit of using keyframes is, that the map points and observations that are connected to certain keyframes can be easily updated when the MCS revisits a place that changed.

In the following, we detail some related work.
For a more detailed overview of available methods and features the reader is referred to \cite{garcia2015vision,gauglitz2011evaluation,CadenaCCLSN0L16}.
%\subsection{Keyframe-Based Approaches}
\section{State-of-the-Art}
\label{ch:tracking:sec:relwork}

%\label{ch:tracking:sec:relwork:subsec:keyframebasedtracking}
In recent years, plenty of keyframe-based SLAM systems were proposed and many of them are publicly available.
Probably the first real-time SLAM system that was based on performing local and global BA over keyframes is PTAM, by Klein and Murray\cite{klein2007parallel}.
One key to success was to split the tracking and mapping components of the system into separate threads running in parallel on a dual-core CPU.
In this way, the mapping thread decides which camera poses to keep and store as keyframes and performs local and global BA asynchronously to the tracking thread.
Latter runs at frame rate and performs matching of map point and camera pose estimation.
Subsequently, PTAM was improved by adding edge features \cite{klein2008improving} and, with the increase in computing power, was implemented on a mobile phone \cite{klein2009parallel}.
%They initialize the map using the 5-Point algorithm \cite{stewenius2006recent}.
The use of image patches as features and the lack of loop closing mechanisms already suggests that large scale operation could be an issue using PTAM as storing, updating and indexing of image patches is costly.
In addition, the global BA is performed over the entire scene, limiting its applicability to smaller workspaces.

To extend the range of PTAM, the authors of \cite{strasdat2010scale} propose a so-called scale-drift aware SLAM system.
First, the keyframe decisions and feature initialization is moved to the tracking thread.
Despite having a higher computational burden, the tracking becomes more stable, as features and keyframes are not initialized asynchronously and are immediately available for pose estimation.
Due to the incremental nature of SLAM, the trajectory estimates start to drift over time, i.e. if the camera visits the same place twice, it will have a different exterior orientation and the reprojected map points exhibit large residuals to the measured image features.
Still, if one is able to automatically detect similar places using place recognition methods \cite{garcia2015vision}, loop closing can be performed.
The loop closing mechanism in \cite{strasdat2010scale} is based on SURF features and a dense surface model.
Then, the trajectory is corrected and optimized over similarity transformations, i.e. also correcting scale-drift.

\begin{figure*}
	\centering
	%trim=l b r t
	\includegraphics[trim = 5mm 125mm 187mm 7mm, clip=true, width=\textwidth]{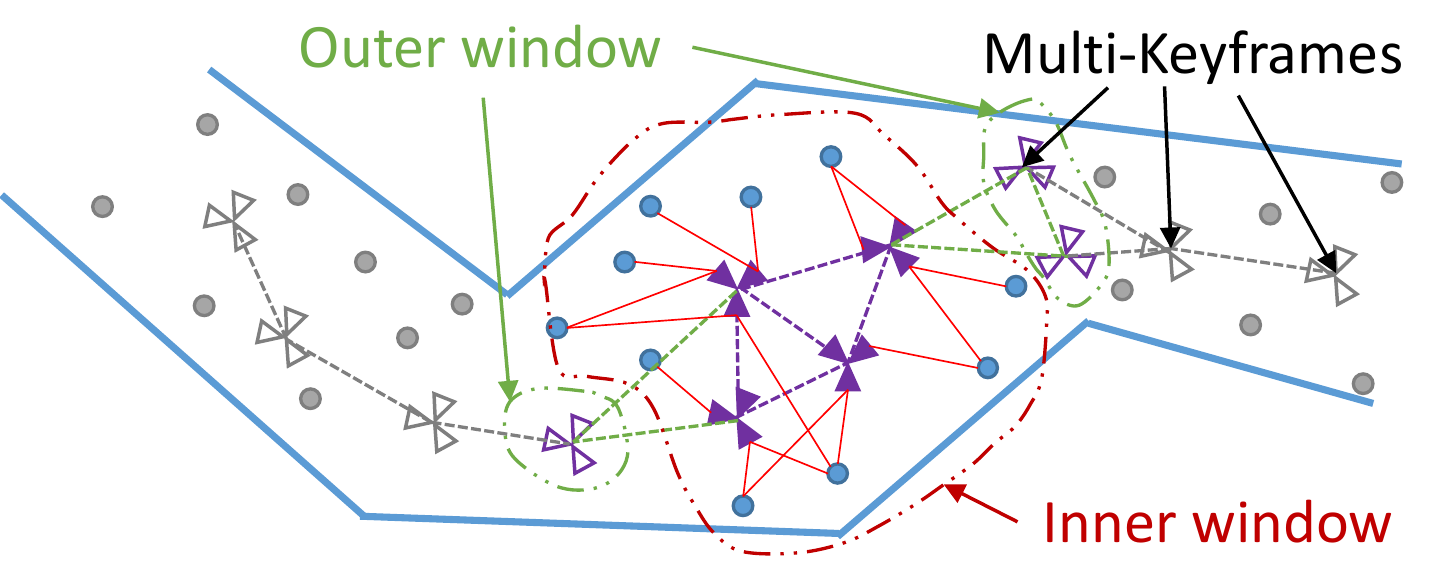}
	\caption{Principle of double window SLAM. }
	\label{fig:doubleWindow}
\end{figure*}

Seminal work \cite{strasdat2011double} focused on keyframe optimization, selection and constraints.
\SeeFig{fig:doubleWindow} depicts the double window approach.
The inner window of active keyframes models the local area.
Poses and map points are optimized using local BA.
The outer window stabilizes the inner window and connects it to the rest of the trajectory.
The question remains how to determine the connectivity between keyframes.
In \cite{strasdat2011double}, the \textit{co-visibility} is introduced.
Instead of creating connections between keyframes based on geodesic or euclidean distance or temporal constraints, image features are used.
By projecting map points to adjacent keyframes and matching their corresponding descriptors, a co-visibility weight can be calculated, that expresses how many equal features are visible in both keyframes.
Apart from being able to update the connectivity if the scene changes, occlusions can be handled a lot better. 

In \cite{pirker2011cd}, a complete SLAM system was proposed including loop detection, re-localization and handling of long-term dynamics.
Co-visibility and local map querying was not only constrained by image to image matches, but also on the level of map points.
A Histogram of Cameras (HoC) descriptor \cite{pirker2010histogram} is used to determine which map points are visible from the current camera pose.
SIFT features are used and extraction and matching is performed on the GPU to achieve real-time performance, which limits its applicability on mobile platforms having only limited computing power.

In \cite{lim2014real}, the authors also build on the double-window optimization and co-visibility ideas but use the FAST detector coupled with BRIEF to exploit the performance of binary features.
Using BRIEF and FAST decreases the time for feature extraction and matching significantly, however, this detector descriptor combination is not rotation and scale invariant and thus limited to in-plane trajectories (like driving cars).
In addition, points are only queried and tracked from the last keyframe.
Thus, the local map structure is not fully exploited.
ORB-SLAM \cite{mur2014orb,murAcceptedTRO2015,orbslam2} is the latest state-of-the-art feature- and keyframe-based SLAM system, that is available on-line.
As the name suggests, ORB features are used, being rotation and scale invariant to some degree.
The map is reused and queried efficiently using co-visibility computed on ORB features.
Place recognition is based on a binary BoW method \cite{Galvez-Lopez2012} and a new map initialization heuristic is introduced, that dynamically switches between fundamental and homography estimation. 
Robust keyframe and map-point culling heuristics ensure a high map quality.
We will build our multi-fisheye camera SLAM upon ORB-SLAM and explain all changes in the next sections. 

The methods described so far either use a single camera (Monocular SLAM) or a stereo configuration (Stereo SLAM). 
CoSLAM (Collaborative SLAM) \cite{zou2013coslam} aims at combining the maps build by multiple cameras moving around in dynamic environments independently.
The authors introduce inter-camera tracking and mapping and methods to distinguish static background points and dynamically moving foreground objects.
In \cite{heng2014self}, four cameras are rigidly coupled on a MAV.
Two cameras are paired in a stereo configuration respectively and self-calibrated to an IMU on-line.
The mapping pipeline is similar to ORB-SLAM and also uses ORB descriptors for map point assignment.
Additionally, the authors propose a novel 3-Point algorithm to estimate the relative motion of the MAV including IMU measurements.
Most recent work on multi-camera SLAM is dubbed MC-PTAM (Multi-Camera PTAM) \cite{harmat2012parallel,harmat2015multi} and is build upon PTAM.
In a first step \cite{harmat2012parallel}, the authors changed the perspective camera model to the generic polynomial model that is also used in this paper. 
This induces further changes, e.g. relating the epipolar correspondence search that now has to be performed on great circles on the unit sphere instead of point to line distances in the plane.
In addition, significant changes concerning the tracking and mapping pipeline had to be made to include multiple rigidly coupled cameras.
Keyframes are extended to MKFs as they now hold more than one camera.
As PTAM, their system uses patches as image features and warps them prior to matching.
Still, the system lacks a mapping pipeline that is capable to perform in large-scale environments.
Subsequent work \cite{harmat2015multi} improved upon \cite{harmat2012parallel} and is partly similar to the SLAM system developed in this thesis in that it uses the same camera model and g2o to perform graph optimization.
On top, the authors integrated an automated calibration pipeline to estimate the relative orientation of each camera in the MCS.
Still the system uses the relatively simple mapping back-end of PTAM instead of double-window optimization that is used in this thesis and has proven to be superior.
In addition, image patches are used as features making place recognition, loop closing and the exploration and storage of large environments critical.

Thus far, all approaches were based on local point image features.
Hence, the reconstructed environment will stay relatively sparse even if hundreds of features are extracted in each keyframe.
This makes it difficult for autonomous vehicles or robots that explore their surrounding to automatically analyze and extract object structure or texture information.
Thus, most of the time, camera localization is coupled with laser scanners \cite{lin2012mapping}, structured light \cite{kerl2013dense}, yielding structured object information.
Recent work on semi-dense \cite{forster2014svo,engel2014lsd} and dense \cite{newcombe2011dtam,concha2015dpptam} camera-based SLAM systems make use of a single camera to estimate dense scene structure instead of reconstructing only point features.

LSD-SLAM \cite{engel2014lsd} is a semi-dense approach that runs on a single CPU in real-time, in contrast to dense methods \cite{newcombe2011dtam} that need heavy GPU support.
Using direct image-alignment by minimizing the photometric error between image discontinuities, the method skips the costly feature extraction and matching stage of all feature-based SLAM systems.
The time saved compensates for the increased BA runtime, as a huge number of observations is included.
In addition, all scale-drift aware loop closing and large scale double window optimizations are included, making LSD-SLAM a state-of-the-art approach that also runs in real-time.
However, loop closing uses FAB-MAP \cite{cummins_newman_2010_icml} for place recognition and thus requires SURF features to be extracted.
Subsequent work extended the method to mobile phones \cite{schops2014semi}, stereo \cite{engel2015large} as well as omnidirectional cameras \cite{caruso2015_omni_lsdslam}.
Instead of coupling camera pose estimation and semi-dense mapping, in \cite{mur2015probabilistic} a semi-dense extension to ORB-SLAM is presented. The semi-dense map is reconstructed from feature-based keyframes using depth consistency tests and a novel correspondence search.
The semi-dense reconstruction is not obtained in real-time but is calculated in a CPU thread running in parallel to tracking and mapping.
The methods yields superior performance compared to LSD-SLAM and it seems that the decoupling is advantageous, especially in dynamic scenes.

\section{Contributions}
We will extend the state-of-the-art ORB-SLAM to multi-fisheye camera systems using MultiCol \cite{UrbanMultiCol2016}.
Our contributions to ORB-SLAM (and ORB-SLAM2 respectively) are the following:
\begin{enumerate}
	\item The introduction of Multi-Keyframes (MKFs).
	\item A hyper-graph formulation of MultiCol.
	\item Multi-Camera loop closing.
	\item Minimal non-central absolute pose estimation methods for re-localization \cite{kneip2013npnp}.
	\item Different initialization method, based on the essential matrix.
	\item Several performance improvements.
\end{enumerate}
In order to use MultiCol, the concept of Multi-Keyframes (MKFs) is introduced. 
Employing a generic camera model \cite{scaramuzza2006b} allows to couple arbitrary central cameras to the MCS.
Instead of employing image patches as features (cf. MC-PTAM), compact binary descriptors proved to be the state-of-the-art when it comes to efficient large-scale re-localization, tracking and loop closing.
\section{Framework}
\label{ch:tracking:sec:framework}
In this chapter, the proposed SLAM system is introduced.
As mentioned previously, the basic structure of our system is build upon ORB-SLAM \cite{mur2014orb,orbslam2}.
The proposed tracking and mapping system is dubbed MultiCol-SLAM.
\SeeFig{fig:MKS_SLAM_OVERVIEW} depicts an overview of the system.
In general it is divided into multiple threads running in parallel and taking care of different aspects.
For the sake of clarity, the loop detection thread is omitted in this figure.
Each adaption to the original ORB-SLAM system is highlighted in red and will be explained in more detail in the following.
Two of the most profound adjustments in MultiCol-SLAM compared to ORB-SLAM are the introduction of Multi-Keyframes (MKFs), i.e. a keyframe consists of multiple images and the use of fisheye cameras.
Both novelties involve some significant changes to the basic design, e.g. bundle adjustment, pose estimation, map point triangulation and relative orientation computation.

With every new set of incoming camera images, the tracking thread extracts point features from every image.
Then, they are stored in a continuous vector, that will later be used to identify and match points across MKFs and mask outliers.
To ensure a fast indexing and querying of feature to camera mappings, we use hash maps (unordered maps in C++) that provide constant time $O(1)$ search.
Like ORB-SLAM, we use the relative orientation between the last two frames to predict the current position of the system.
The local map points are projected to the MCS and matched to the extracted features from the current frame.
If enough matches are retained from the set of putative correspondences after an initial robust pose optimization using MultiCol, the tracking thread starts to search for more matches, assigns the reference MKF and decides if a new MKF should be added and passed over to the mapping thread.
If the initial pose estimation fails, GP3P \cite{kneip2013npnp} and RANSAC are used to perform re-localization of the MKS using the map points assigned to a set of recent MKFs.
This is different compared to ORB-SLAM where a single camera and the non-minimal PnP solver EPnP \cite{lepetit2009epnp} is used in a RANSAC loop to find potential map point matches and estimate the current camera pose after tracking failure.
The tracking thread is detailed in \SeeSec{ch:tracking:sec:trackingthread}.

Each time the tracking thread passes a new MKF to the mapping thread, recently created map points that do not fulfill certain conditions are deleted from the map.
Then, new map points are triangulated over MKFs that are in the vicinity of the added MKF.
Here, the vicinity is determined by the co-visibility graph.
In contrast to ORB-SLAM, where features are only triangulated between images of the same camera, the reconstruction is now performed over images of different cameras as well.
Subsequently, local bundle adjustment is performed to adjust the poses of the MKFs that are part of the local map, as well as all map points.
In addition, the mapping thread decides which MKFs are redundant and deletes them from the map.
The mapping thread is detailed in \SeeSec{ch:tracking:sec:mappingthread}.

The loop closing thread searches for potential loop closures with every new MKF that is added.
To decide if a place was already visited before and to identify MKFs as loop candidates, the system uses a Bag-of-Binary-Words framework \cite{Galvez-Lopez2012}.
If a loop is detected, the essential graph (a sparse version of the co-visibility graph) is optimized.
To correct for the scale-drift, the optimization is carried out over similarity transformations that connect the MKFs.

\begin{figure*}
	\centering
	%trim=l b r t
	\includegraphics[trim = 5mm 110mm 70mm 5mm, clip=true, width=\textwidth]{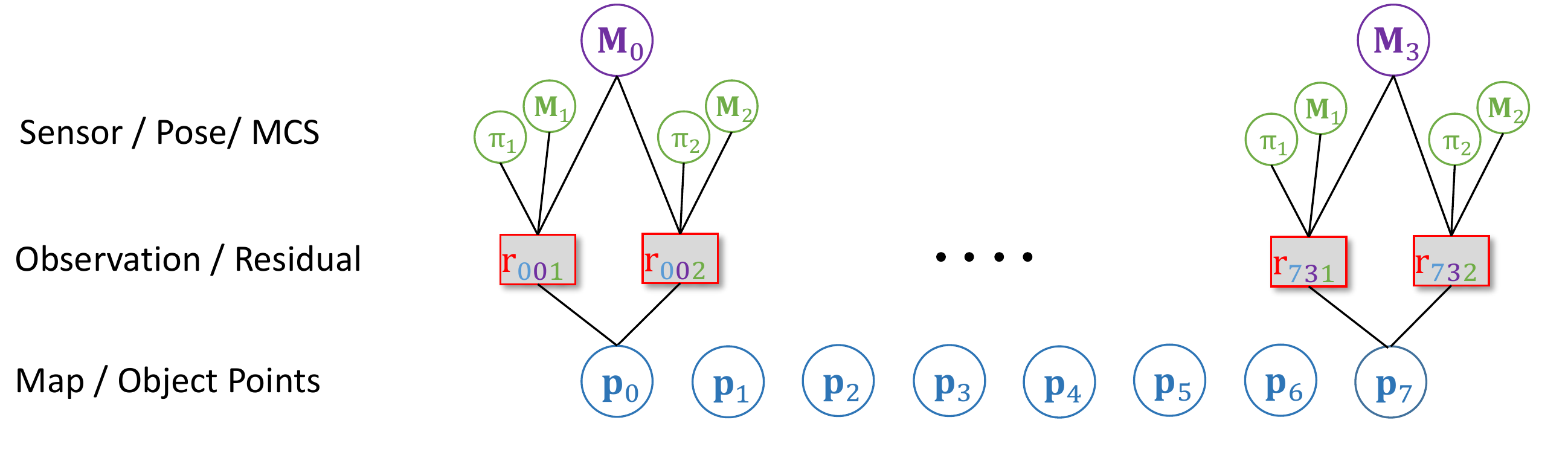}
	\caption{Depicted is the hyper graph model of MultiCol. Parameters are denoted as circles (vertices) and measurements as boxes (edges). In such a hyper graph edges can be connected to multiple vertices. In this example the $i=1,..,7$ map points $p_i$ are observed by a MCS at a particular time from pose $\mathbf{M}_t$, with $t=1..3$. The MCS consists of c=1..2 cameras that have a relative orientation $\mathbf{M}_c$ w.r.t the body frame and an interior orientation $\pi_c$.}
	\label{fig:hypergraphMultiCol}
\end{figure*}

Like ORB-SLAM, we use the graph optimization framework g2o \cite{kummerle2011g} for all optimizations.
The difference to ORB-SLAM is how we model the tracking and mapping pipeline.
As multiple cameras observe the scene from each pose (\SeeFig{fig:SlamProblem}) and also one map point can be observed by multiple cameras at the same time, the graph can not be represented by binary edges anymore (edges that connect two vertexes).
Instead, we extend the graph to a hyper graph that we used to model MultiCol (cf. \SeeFig{fig:hypergraphMultiCol}) where edges can connect to an arbitrary number of vertexes.

\subsection{The MultiCol Model}
In this short subsection, we will briefly recapitulate the MultiCol model.
For a more in-depth introduction to MultiCol and the involved camera model the reader is referred to \cite{urban2015improved,UrbanMultiCol2016}.

\begin{figure}[h]
	\centering
	\includegraphics[trim = 10mm 110mm 240mm 15mm, clip=true,width=\columnwidth]{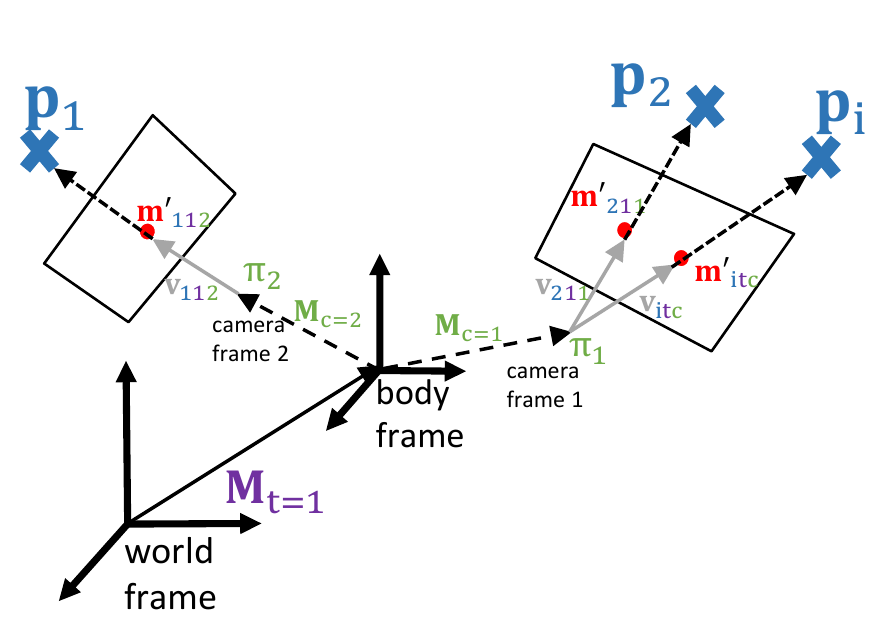}
	\caption{Depicted is the body frame concept and all involved parameters.}
\end{figure}
Given a homogeneous transformation matrix $\mathbf{M}$ the transformation and projection of a world point $\mathbf{p}_i$ to its corresponding image point $\mathbf{m}_{itc}$ in camera $c$ at time $t$ is given by:
\begin{equation}
\boxed{\mathbf{m}_{itc} 
= 
\pi^g_{c}(\mathbf{p}_{itc}) 
= 
\pi^g_{c}(\mathbf{M}_c^{-1}\mathbf{M}_t^{-1}\mathbf{p}_i)}
\label{eq:MultiColModel}
\end{equation}
where $\pi^g_{c}$ is the generic camera model presented in \cite{scaramuzza2006b,urban2015improved} modeling all types of central cameras.
If only perspective cameras are used, this could be exchanged with the calibration matrix $\mathbf{K}$ and some additional distortion coefficients.

In the next section, the basic entities and methods that are used to represent the environment, i.e. map points, Multi-Keyframes and the co-visibility graph are detailed.

\subsection{Map Entities}
\label{ch:tracking:sec:mapentities}
\begin{figure*}
	\centering
	\includegraphics[trim = 1mm 45mm 85mm 1mm, clip=true, width=\textwidth]{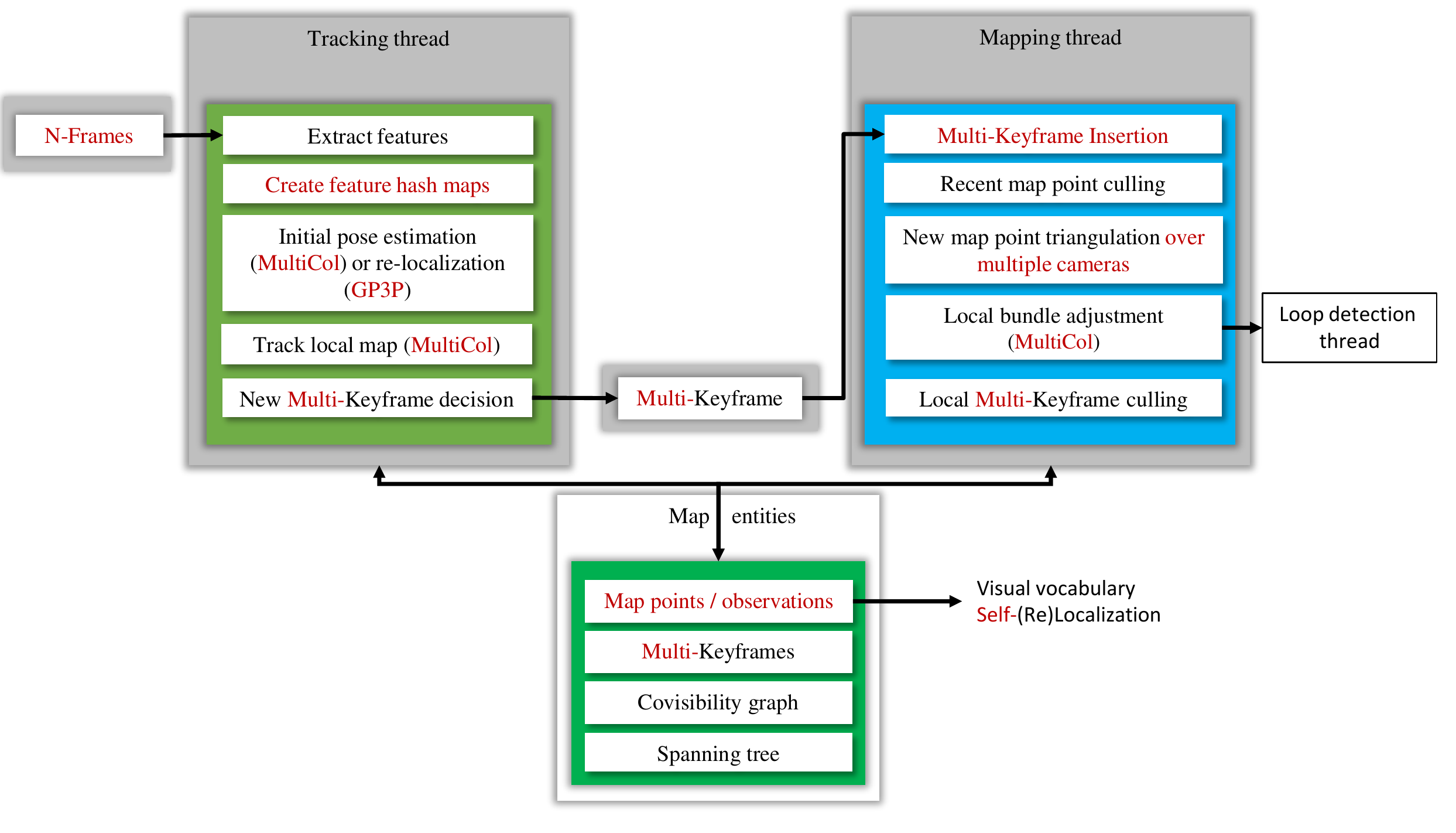}
	\caption{Depicts the MultiCol-SLAM framework without the loop closing thread. Red text depicts the modules where significant difference between our method and ORB-SLAM exist.}
	\label{fig:MKS_SLAM_OVERVIEW}
\end{figure*}

\paragraph{Map Points}
The map point is the most basic entity of the framework.
Each map point $\mathbf{p}_i$ has the following attributes, properties and variables:
\begin{itemize}
\item The 3D position $\mathbf{p}_i = [X_i,Y_i,Z_i]^T$ in world coordinates.
\item A maximum $d_{max}$ and minimum $d_{min}$ distance at which the point can be observed. This distance is used to reduce the number of points that are queried for local map tracking.
\item The viewing direction $\mathbf{n}_i = [n_x,n_y,n_z]^T$. 
\end{itemize}

\paragraph{Multi-Keyframes}
In contrast to ORB-SLAM, each keyframe stores multiple images and is thus called Multi-Keyframe.
Again, each Multi-Keyframe $\text{MKF}_t$ created at time $t$ has a number of attributes and variables:
\begin{itemize}
\item The MCS pose $\mathbf{M}_t$. 
\item A MCS object that stores the intrinsics of each involved camera and the extrinsics ($\mathbf{M}_c$) of the MCS.
This object also performs the forward and back projection of world and image points.
\item All features that are extracted from each camera. The features are stored in continuous vectors and thus a fast feature to camera search is needed.
For each image point, we store two representations.
One is its 2D image coordinate $\mathbf{m}'$ that we use extensively in MultiCol bundle adjustment and pose estimation.
In addition, we store the corresponding 3D bearing vector $\mathbf{v}$.
Latter will be used in various geometry related algorithms, e.g. essential matrix estimation, epipolar search and absolute pose estimation (GP3P).
\item The Bag-of-Binary-Words representation.
\end{itemize}
\paragraph{Co-Visibility Graph}
As in ORB-SLAM, the co-visibility graph is represented as a weighted undirected acyclic graph.
The weight $\chi$ of each edge that connects two nodes (MKFs) in the graph is calculated as the number of map points the two MKFs share.
In \cite{strasdat2011double}, a minimum weight $\chi_{min}$ between 15 and 30 was used to insert a connection.
In contrast, the authors of ORB-SLAM do not impose a constraint on the minimal weight. 
When needed, the co-visibility graph is queried with a threshold, to only return nodes with a weight larger than $\chi_{min}$, but the connectivity is kept very dense.
\paragraph{Map Initialization}
The initialization of the system comprises the estimation of an initial map and the corresponding camera poses.
In general, the initialization of the map is the first crucial step in camera based SLAM system.
The accuracy and robustness of the initial reconstruction has a significant impact on the overall performance of the system.

The authors of ORB-SLAM introduce an algorithm that switches between scene reconstruction using the fundamental matrix $\mathbf{F}$ or estimating a homography $\mathbf{H}$ using a heuristic.
In general, this is a challenging task and often ignored, assuming the first camera motion introduces enough parallax.
If a homography describes the current scene better, i.e. if the camera is only rotating or observes a completely planar scene, initialization is suppressed.
If a proper fundamental matrix is found, the scene is initialized, by reconstructing camera poses and scene points, followed by bundle adjustment eliminating the gauge freedom by fixing the first camera.

With MultiCol-SLAM, two issues arise that require some adaption.
On the one hand, both $\mathbf{F}$ and $\mathbf{H}$ matrices contain the perspective camera matrix $\mathbf{K}$. 
For omnidirectional or fisheye cameras and especially the camera model employed in this work, a camera matrix does not exist.
Still, if images points would be undistorted to their corresponding location in a perspective image, an application would be possible, although points at the image border would be lost.
On the other hand, we are actually looking for the relative motion between two MCSs.
Thus a map has to be initialized for each camera separately and then fused somehow.
A different approach is to directly estimate the relative orientation between two MCS poses which is equivalent to computing the relative pose between two generalized cameras \cite{yu2004general}.
We experimented with the linear 17-pt \cite{yu2004general}, a 6-pt \cite{stewenius2005solutions} and a 8-pt algorithm \cite{kneip2014efficient} all part of OpenGV. 
The two polynomial solvers are relatively slow and the linear 17-pt algorithms is numerically very unstable.
Recently, a new method was published \cite{ventura2015efficient} but we leave the investigation for future work.

As the initial reconstruction of the SLAM trajectory is not at the core of this work, we propose a rather practical than generic methodology.
We estimate an essential matrix $\mathbf{E}$ in a RANSAC loop between the same camera from different MCS poses and choose the one with the most inliers and the highest translational magnitude.
Then, we exploit a slight overlap between the FoVs and search for the map point projection in all other cameras.
Finally, we perform bundle adjustment over all observations and the two MCS poses.
This routine, however, only works robustly if small overlap exists.

\subsection{Tracking Thread}
\label{ch:tracking:sec:trackingthread}
In this section, the tracking thread is detailed.
It is the core of our multi-camera tracking system as it handles not only the current state but performs feature extraction, matching and pose estimation.
%To switch between different states, e.g. re-localization, exploration or model-supported tracking, we will use various heuristics.
At the same time, it is the only thread that has to run in real-time, i.e. at frame rate.
If this is not the case, incoming camera images will be dropped and tracking will suffer.
Thus, an efficient implementation of all methods is essential.
In addition, the tracking thread handles the MKF insertion and takes care of distributing work to all other threads.

All optimizations are carried out using iterative re-weighted least squares (IRLS) using Levenberg-Marquardt regularization and a robust Huber kernel.
Huber suggested to calculate the tuning constant $e$ as $e=1.345\sigma$, where $\sigma$ is some estimate of the standard deviation of the residuals, that we set to $\sigma=2$.
%We calculate latter after each pose estimation and each local bundle adjustment respectively, to get an estimate of the current residual mean standard deviation.
%Thus the Huber tuning constant is adapted after each iteration of the corresponding optimization.
After each optimization, outlier edges (measurements) are found and eliminated by testing the residual against the Huber value. 

\paragraph{Feature extraction}
The standard ORB detector extracts FAST corners at multiple scale levels (usually 8) and retains a certain number of corners per level that fulfill the Harris cornerness measure.
In ORB-SLAM each image is additionally divided into several cells on each pyramid level.
Then, the extractor tries to find at least 5 corners per cell to ensure a homogeneous distribution of feature points in the image.
If this is not the case the cornerness threshold is adapted.
Finally, the feature orientation and a ORB descriptor is computed.

As FAST corners usually need a re-training in new environments, we chose to utilize AGAST \cite{Mair2010c} corners instead.
Note that, in theory, none of the efficient corner extractors is suited for highly distorted (fisheye) images.
To achieve a higher repeatability, the pixel positions of the Bresenham circles used for intensity testing would have to be interpolated, hence completely destroying the efficiency.
Some detectors for omnidirectional images were presented in \cite{hansen2008spherical,lourencco2012srd}.
These methods report high repeatability scores, however, are to slow for real-time applications.
An interesting adaption of FAST and ORB to spherical images was proposed in \cite{zhao2015sphorb} called SPHORB, however, without providing the corresponding source code.

In MultiCol-SLAM, we extract ORB descriptors on each AGAST feature that is extracted in different cells over multiple scale levels keeping in mind that neither the detector nor the descriptor is particularly suited for distorted images.
It shows, however, that the robust tracking and mapping back-end as well as the restriction of feature matching to local image areas (guided search) is able to compensate for the drawbacks and and weaknesses of the feature extraction stage.

In addition, we extend the feature extraction module of the implementation to work with  detectors and descriptors that are part of OpenCV.

\paragraph{Tracking from the Previous Pose}
This step is similar to ORB-SLAM.
First, the current pose is estimated by using a constant velocity motion model.
Then, local map points assigned to the last pose are projected to each camera of the current MCS and a guided search is performed around the projected location.
With this initial set of matches, the MCS pose is optimized using MultiCol on fixed map points and outlier measurements are marked by identifying edges with residuals over $e$.
With the optimized pose, the guided search is repeated, to identify more potential matches and the pose is optimized again.

\paragraph{Re-Localization}
As soon as the tracking thread indicates a tracking failure, re-localization is carried out.
This happens if not enough points are retained after the initial pose tracking.
Then, the images are converted into their corresponding Bag-of-Words representation, and the recognition database is queried for potential MKF candidates.
We iterate through each MKF and match all associated map points to the keypoints detected in the current frame yielding a set of putative correspondences.
In ORB-SLAM, the initial pose estimate is found using EPnP in a RANSAC loop.

We exchange EPnP for two reasons.
On the one hand, EPnP requires more than three points to estimate the current pose of the camera and is thus not a minimal solution.
The EPnP estimate is furthermore rather unstable for only few points \cite{urban2016MLPNP}.
On the other hand, as we build our system on a MCS, we try to solve for the six degrees of freedom of the MCS pose using observations from multiple cameras.
Thus, GP3P+RANSAC \cite{kneip2013npnp} is used to find a putative set of inliers for all MKF candidates.
If enough inliers are retained and RANSAC did not exceed a predefined number of iterations, we refine the initial pose estimate obtained by GP3P for each MKF over all inliers using UPnP \cite{kneip2013npnp}.
Finally, the pose is optimized with MultiCol, again suppressing map point correspondences with high residuals.
If more than a predefined number of points (we set this to 15) is retained after final pose optimization, re-localization was successful and the tracking thread goes back into its usual behaviour. 

\paragraph{Tracking the Local Map}
The final step of the tracking loop is important, as it reinforces the visual connectivity and builds a densely connected co-visibility graph.
Having estimated an initial pose either from tracking the previous pose or after re-localization, the local map is projected to the MCS to find more matches.
To identify which map points are contained in the local map, a reference MKF to the current pose has to be found first.
Thus, we take the list of map points that are currently assigned to the MCS pose entity and from which the current pose was estimated.
As each map point stores a list of MKFs it has been observed in, we can then iterate through all map points and count their occurrences.
Care has to be taken, as each map point can be observed multiple times from each MKF.
The MKF with the most occurrences is taken as the reference MKF and a set of local MKFs is queried from the co-visibility graph.

Then, the following steps are performed consecutively over each point $\mathbf{p}_i$ in the local map:
\begin{itemize}
	\item[(1)] Project $\mathbf{p}_i$ into each camera of the MKF. Discard if projection is not inside the mirror boundary of a specific camera. Otherwise add the point to potential candidates.
	\item[(2)] Compute the angle between the current bearing vector $\mathbf{v}_i$ and the map point viewing direction $\mathbf{n}_i$ and discard if angle is larger than $50^{\circ}$.
	\item[(3)] Compute the distance $d_i$ from the current MCS pose to the map point.
	If this distance is outside the interval $d_{min}$ and $d_{max}$ that are defined by the scale invariance region of the image pyramid, discard the point.
	\item[(4)] Get all descriptors around the projected location of the map point at a given scale and match them to the map point descriptor.% using the unilateral or bilateral Hamming distance if available.
\end{itemize}
Finally, the pose is optimized for the last time. 
If not enough matches are retained, the tracking thread goes into re-localization mode.

\paragraph{New Multi-Keyframe Decision}
From a robustly estimated MKF pose that is tightly connected to the local map and co-visible MKFs the tracking thread decides if it is time to add a new MKF to the map.
The insertion takes place if the following conditions are met.
All thresholds are set according to the FPS of the camera system (our MCS runs at $\text{FPS}=25)$:
\begin{itemize}
	\item[(1)] More than $0.5 \cdot \text{FPS}$ frames have passed from last MKF insertion and the local mapping thread is idle.
	\item[(2)] A certain amount of poses must be successfully tracked from the last re-localization. In our case this threshold is set to the current frame rate.
	\item[(3)] At least 50 points are tracked from the current MCS pose.
	\item[(4)] Less than 90\% of the current map points are assigned to the reference MKF. Thus, MKFs are only inserted if the visual change is big enough.
%	\item[(5)] A minimum distance between the current MCS pose and the reference must be exceeded. We roughly estimate this distance from the median scene depth $\delta_{med}$ and apply it after a couple of MKFs are already inserted into the map. Given the basis $b$ between two cameras, their focal lengths $c$ in meter and an estimate of the image measurement accuracy $\sigma_{m}$ the reconstruction error in z direction $\sigma_{z}$ can be approximated by $\sigma_{z}=\frac{\delta^2\sigma_{m}}{bc}$.
%	Thus the minimum distance between two MKFs can be roughly derived from a desired reconstruction quality. Usually we set $\sigma_{z}=0.1m$.
\end{itemize}
%In ORB-SLAM, the last condition is avoided. 
%For MultiCol-SLAM where a 360$^\circ$ view of the environment is present at all time, we found it to be important, as to many MKFs where inserted and the reconstruction quality suffered.
\subsection{Mapping Thread}
\label{ch:tracking:sec:mappingthread}
This section details the mapping thread (cf. \SeeFig{fig:MKS_SLAM_OVERVIEW}).
Asynchronously to the tracking thread, it extends the map by triangulating new points, performs local bundle adjustment and deletes redundant map points and MKFs from the map.
Every time it finishes one loop, the entities are fed back to the tracking thread and become available.

\paragraph{Multi-Keyframe Insertion}
As soon as the tracking thread decides to insert a new MKF into the map, the mapping thread starts to update the co-visibility graph.
The last step in the tracking thread ensured a tight connectivity.
Thus a new node is added to the graph and the edge weights are updated with the number of map points each MKF shares with the new MKF.
In addition, a Bag-of-Words representation of the new MKF is computed and saved in the recognition database.

%If the tracking takes places inside a 3D model, the following steps are performed:
%\begin{itemize}
%	\item[(1)] Render the depth images from the current MKF pose $\mathbf{M}_t$, that was estimated from tracking feature points.
%	\item[(2)] Extract image edges. For each image of the MCS, we first extract gradients using a 3$\times$3 Sobel kernel.
%	Subsequently, the derivatives in x and y direction are passed to a Canny edge detector \cite{canny1986computational} to get thin edges.
%	For all remaining edge pixels, we then extract edge orientations.
%	\item[(3)] Project all model edge and companion points to the MCS and perform a simple depth test using the rendered depth image. We add a small offset of 5 cm to the depth values, in order to compensate for uncertainties in the MCS pose. For very large 3D models, the model edge points could be queried from a kdtree with a radius search, to decrease the number of projection and test calculations.
%	\item[(4)] Estimate the orientation $\phi_e$ of the projected model edges that passed the depth test in the image using their projections $\mathbf{m}'_{e}$ and $\mathbf{m}'_{co}$. 
%	\item[(5)] Search perpendicular to $\phi_e$ for image edges in both directions for $t$ pixels. An image edge is found if the orientation difference is below some threshold $t_{e}$. Again, we compare edge orientations using the cosine similarity, to avoid polarity issues. As soon as an edge is found in one direction, we stop the search in that direction.
%\end{itemize}
\paragraph{Recent Map Point Deletion}
The mapping threads stores a list of recently added map points from the last three MKFs.
All map points under consideration have to pass the following conditions to remain in the map.
The conditions do not only take visibility in subsequent MKFs into account but also the appearance between them.
\begin{itemize}
	\item[(1)] A map point has to be found in at least 25\% of its predicted MCS poses.
	\item[(2)] The map point must be observed from at least three MKFs.
%	\item[(3)] If the mean accuracy (from $\Sigma_{\mathbf{p}_i\mathbf{p}_i}$) after reconstruction from three MKFs has to be below some threshold $\tau_{\Sigma}$.
\end{itemize}
As in ORB-SLAM, a map point can only be removed from the map if it is visible in less than three MKFs.
This can happen if MKFs are removed from the map.
%\paragraph{Adjust MKF to Model}
%If tracking is performed inside a 3D model, we adjust the MKF to the model.
%After MKF insertion, we have pairs of model edge points $\mathbf{p}_e$ and corresponding image measurements $\mathbf{m}'_{eI}$. 
%Thus, we have narrowed down the problem of aligning the MKF to the model edges to pose optimization.
%After adjusting the MKF pose, we re-render the model.
%This time, not only the depth, but also normal and world coordinate images are saved.

\paragraph{New Map Point Triangulation over Multiple Cameras}
Map points are created by triangulation from MKFs. 
First the co-visibility graph is queried for five MKFs from the current reference MKF.
Then the following iteration is carried out to create new map points:
\begin{itemize}
	\item[(1)] Check if the baseline between the reference and the current neighboring MKF is big enough.
	\item[(2)] Search image points for triangulation. First the Essential matrix $\mathbf{E}$ is calculated for each camera pair. E.g. in this case of three cameras, we get nine essential matrices in two MKFs. Those will be used to verify that matched points lie on the corresponding epipolar great circle.
	\item[(3)] Take all descriptors from the reference MKF and match them to all descriptors from the queried MKF. 
	\item[(4)] Discard, if point is too far from epipolar great circle or Hamming distance is above threshold.
\end{itemize}
%If model-supported tracking is performed, we additionally do the following:
%\begin{itemize}
%	\item[(5)] Get the world coordinate images of each MKF and extract two world points $\mathbf{p}_{MKF_i}$ and $\mathbf{p}_{MKF_j}$ for each match.
%	If the distance between both world points is below a threshold, we directly assign the newly created map point to class $C_m$, set the map point coordinate $\mathbf{p}_i$ to the mean world point coordinate of $\mathbf{p}_{MKF_i}$ and $\mathbf{p}_{MKF_j}$ and extract a normal vector $\mathbf{n}_i$ from the rendered normal images.
%\end{itemize}
Finally map points are triangulated and tested for positive depth, parallax and reprojection error in both cameras from which they were triangulated.
If the newly created map point passes all tests, it is added to the map, a descriptor mask is learned and it is passed to the recently added map point list, that in turn is used to cull bad map points.
\paragraph{Map point fusion} \label{para:mappointfusion}
As the triangulation step might have yielded redundant points that were already in the map, the next step is to fuse those point duplications.
Therefore, the map points connected to the current MKF are first projected to all MKFs that are connected in the co-visibility graph.
Within each MKF the map points are projected to each camera.
Then, all features in a local area around the projected point are queried and matched.
If a match is found, that also lies on the epipolar great circle, the map points are fused.
If the matched image point is not connected to a map point yet, it is added as an observation.
After projecting from the current to all connected MKFs, the same procedure is carried out vice versa.

\paragraph{Local Bundle Adjustment}
The local bundle adjustment optimizes over poses and map points in the inner window (cf. \SeeFig{fig:doubleWindow}).
Therefore, we query the co-visibility graph from the current reference MKF to get a set of MKFs.
Then, all map points that are seen from this set of MKFs are added to the local map.
In addition, the outer window is found, by looping over the current set of local map points and identifying MKFs that are not yet part of the local set of MKFs.
This stabilizes the trajectory and connects it to the rest of the map.

Again the MultiCol equations are used, but this time we optimize over the local map points $\mathbf{p}_i$ and poses $\mathbf{M}_t$.
Although hundreds of points and dozens of poses are subject to optimization, the bundle adjustment problem can be efficiently solved by exploiting the special sparsity structure of the Jacobian and Hessian respectively.
The special structure comes from the fact that only point-pose but no point-point or pose-pose constrains exist. 
The resulting normal equations can the be efficiently solved using the Schur complement trick.
For more details and insights on the subject, the reader is referred to \cite{engels2006bundle}.
Subsequently, the local map is updated and passed back to the tracking thread.

%Care has to be taken if model-supported tracking is enabled.
%Then, the MKF poses have already been adjusted by aligning image and model edges.
%In this case, we only optimize over the local map points $\mathbf{p}_i$.

%\paragraph{Estimate Map Point Class}
%After adjusting the local map points, we estimate their class affiliation.
%Recall that we would like to distinguish between three different classes, i.e. model point $C_m$, stable points inside the model $C_s$, and map points outside the model volume, reconstructed during full SLAM.
%
%A map point is considered to belong to class $C_s$ if it has more than three observations across MKFs and tracking takes place inside a model.
%If the distance between its reconstructed and adjusted world coordinate $\mathbf{p}_i$ and the coordinate extracted from the world coordinate image of the MKF, we set the class to $C_m$.
%The threshold is set to 5cm, i.e. the point has to be closer than 5cm to the model to be classified as model point.
%For larger scenes, a heuristic based on the pose and point uncertainties could be found.

\paragraph{Local Multi-Keyframe Culling}
Like in ORB-SLAM, we delete a MKF from the map if any pair of MKFs shares more than 90\% of the same map points.
Instead of counting map point observations for each camera of the MCS, we count only the occurrence per MKF.

\subsection{Loop Detection and Closing Thread}
\label{ch:tracking:sec:loopclosing}
\begin{figure*}
	\centering
	\includegraphics[trim = 0mm 100mm 200mm 8mm, clip=true, width=\textwidth]{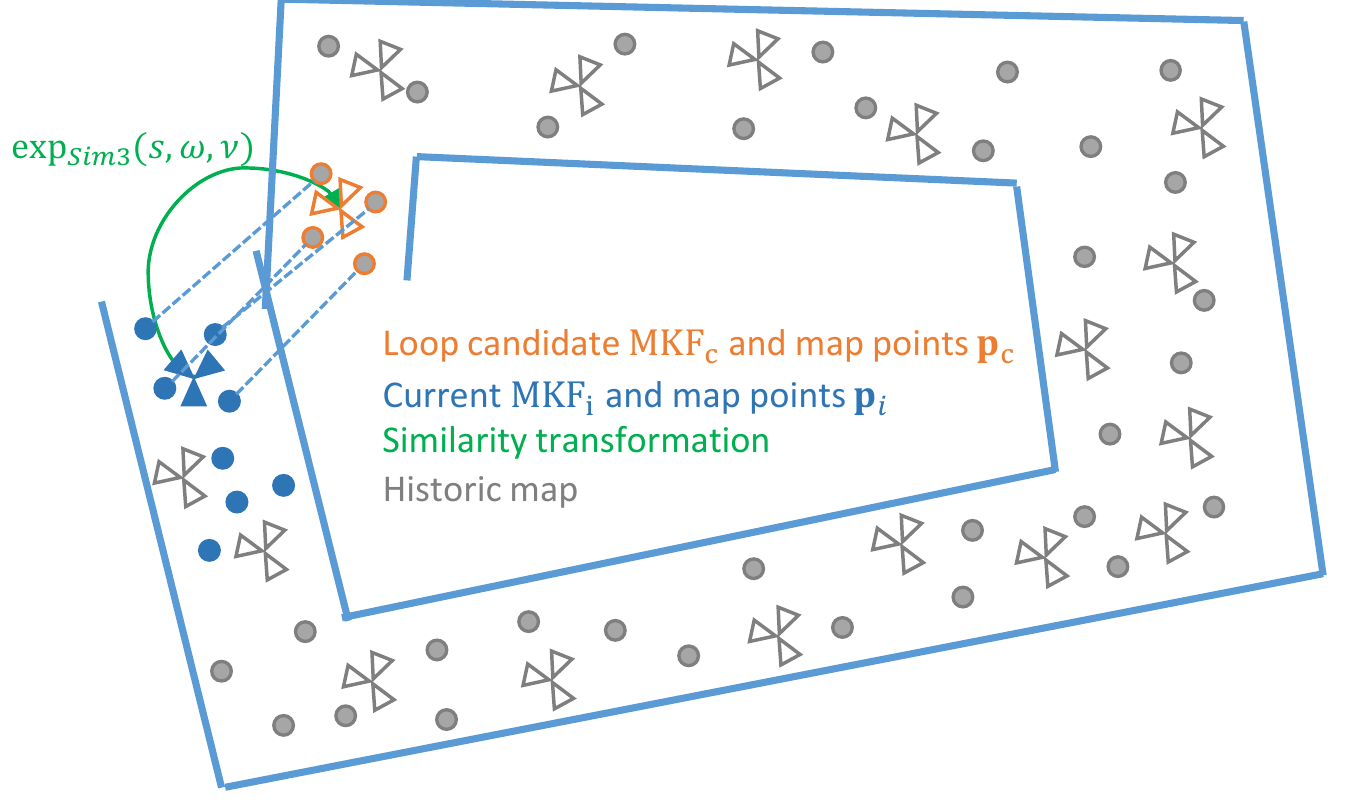}
	\caption{Depicts the loop closing problem. If the SLAM trajectory was estimated without drift, the orange and blue map points should coincide. As this is in general not the case, a similarity transformation can be estimated that aligns both parts of the trajectory over the map points. Then, the alignment error can be used to correct the remaining MKF poses and map points by projecting it back through the map.}
	\label{fig:loopProblem}
\end{figure*}

In this section, the loop detection and correction procedures are detailed.
They are similar to the methodology proposed in \cite{mur2014orb} but extended and adapted to multi-fisheye camera systems.
As measurement errors accumulate over time, the estimated trajectory starts to drift in seven degrees of freedom, i.e. translation, rotation and scale.
This effect is visualized in a toy example in \SeeFig{fig:loopProblem}.
Although the MCS visits the same place, the current local map (depicted in blue) does not coincide with the historic map (depicted in gray and orange).
Although the map does not spatially align with the start of the trajectory, its local map structure is very similar assuming that the system reconstructs a large amount of map points at the same location, i.e. the reconstruction is repeatable.
The goal of the loop detection thread is to identify the situation depicted in \SeeFig{fig:loopProblem}. 
It detects the loop candidates from the historic trajectory and corrects the loop by propagating the loop closing error along the trajectory.

\begin{figure*}
	\centering
	\includegraphics[trim = 0mm 100mm 145mm 1mm, clip=true, width=\textwidth]{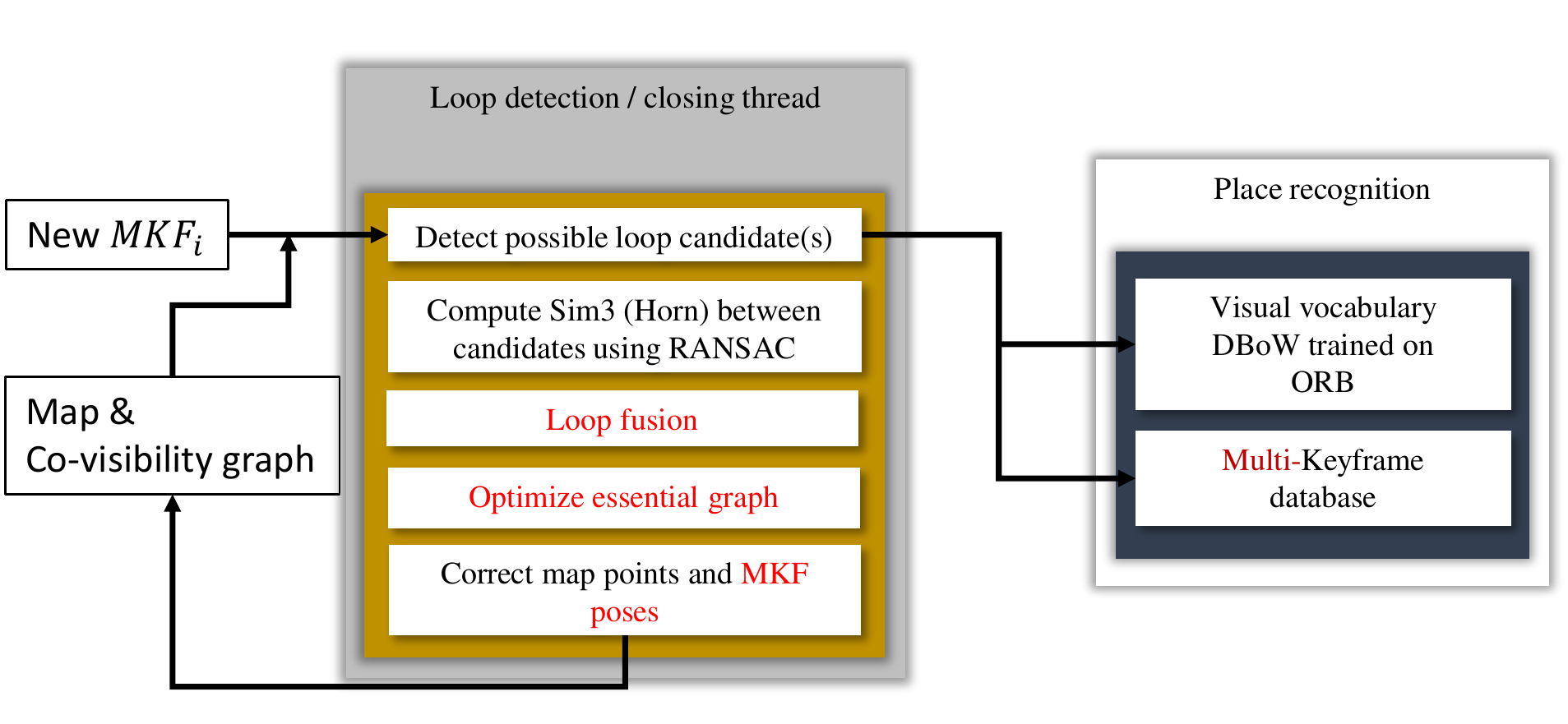}
	\caption{Depicts the loop detection and correction thread. It extends \SeeFig{fig:MKS_SLAM_OVERVIEW}. Each time a new MKF is inserted, the loop detection thread tries to detect possible loop candidates from the MKF database. First, a BoW score $s_{min}$ is computed for all frames that are connected to the new MKF in the co-visibility graph. Then the MKF database is queried and only MKFs with a score higher than $s_min$ are declared as candidates for loop detection. The right part shows the two components of the place recognition class.}
	\label{fig:loopclosing}
\end{figure*}

The loop detection and correction procedure, as well as the place recognition database and visual vocabulary components are depicted in \SeeFig{fig:loopclosing}.
Each step is detailed in the following sections.

\paragraph{Candidate detection}
As soon as the system revisits a place or parts of a scene it has seen and reconstructed before, it should load the associated local map points and start tracking from them instead of starting to reconstruct the scene again.
This, however, is a challenging task, as the local map is solely queried from the co-visibility graph.
One possible way would be to query the map points spatially, i.e. that we  query the nearest neighbor map points in a specified radius from the current MKF.
This method could work for smaller loops, but as soon as larger loops occur, the corresponding local maps could lie dozens of meters apart, which is sketched in \SeeFig{fig:loopProblem}. 
A more favorable solution is to use visual cues to identify possible loop candidates.

After local bundle adjustment, the current multi-keyframe MKF$_i$ is handed over to the loop detection and correction thread.
Now the MKF database could be directly queried for visually similar MKFs. 
However, the number of database results has to be limited somehow.
This could either be achieved by taking a fixed number of MKFs sorted by their similarity score.
Latter, however, is an absolute measure whose magnitude is unknown, e.g. we queried 10 MKFs, but all have a very bad similarity score.
Thus a way of getting a relative measure of the quality of the current similarity score is needed.

Hence, the BoW vectors of all MKFs that are connected in the co-visibility graph of the current MKF$_i$ are queried.
Then, the similarity score between all BoW vectors and the current MKF BoW vector are calculated.
The lowest score $s_{sim}$ is taken as a similarity threshold, i.e. we only take MKFs from the database as candidates if their score is higher than $s_{sim}$ and thus more similar, than the most dissimilar MKF in the co-visibility graph.

To further reduce the number of possible false candidates, MKF candidates are only accepted if they are part of a consistent group of connected MKFs in the co-visibility graph after several consecutive MKF insertions.
Each loop candidate is connected to a number of MKFs (a group) in the co-visibility graph.
A group is accepted to be consistent with the previous group if it they share a MKF.
\paragraph{Transformation estimation}
If one or more candidates are accepted, the similarity transformation between the current MKF$_i$ and all candidates can be estimated.
Lets assume for now, that we have one candidate MKF$_c$.
Looking back at \SeeFig{fig:loopProblem}, the goal is to find the similarity transformation $\mathbf{S}$ between the map points $\mathbf{p}^i$ assigned to MKF$_i$ and the map points $\mathbf{p}^c$ assigned to MKF$_c$:
\begin{equation}
\mathbf{p}^i_{mcs}
=
\mathbf{S}\mathbf{p}^c_{mkf}
=
\begin{bmatrix}
s\mathbf{R} & \mathbf{T} \\
\mathbf{0}^T & 1
\end{bmatrix}
\mathbf{p}^c_{mkf}
\end{equation}
Instead of taking all map points that are connected to both frames, the descriptors assigned to each map point by the MKF are matched in advance.
This leaves us with a subset of possible map point correspondences.
Yet, this set contains outliers that are either caused by wrong descriptor matches or the distance ratios between reconstructed points is too big, caused by bad reconstruction.
Hence, RANSAC is used to find a similarity transformation using Horn's quaternion (\cite{Horn:1988:CFSOM}) method as a model and thus 3D-3D \footnote[1]{Recently, new methods were proposed, that solve the generalized relative orientation and scale problem (similarity between MCSs) using 2D-3D \cite{sweeney2014gdls} or 2D-2D \cite{sweeney2015computing} correspondences only to compute $\mathbf{S}$.
Latter would alleviate the effects of reconstruction error, however, depends on an additional constraint, i.e. the current vertical direction needs to be determined. We leave the integration and investigation to future work.} correspondences.

First the set of map points matches is transformed to the respective MKF:
\begin{eqnarray}
\mathbf{p}^c_{mkf} = \mathbf{M}^c_t \mathbf{p}^c &
\mathbf{p}^i_{mkf} = \mathbf{M}^i_t \mathbf{p}^i
\end{eqnarray}
where $\mathbf{M}_t$ is the respective pose of the MKF.
Obviously, the points can not be transformed to each camera of the MKF, as map points can be observed from multiple cameras and the only common frame is the MKF frame.

Subsequently RANSAC iterations are performed.
Three points are selected from each point cloud, and the transformation is estimated.
To decide whether the transformation is accepted, the map points are transformed from MKF$_c$ to MKF$_i$ and vice versa using the estimated similarity:
\begin{eqnarray}
\hat{\mathbf{p}}^i_{mkf} = \mathbf{S} \mathbf{p}^c_{mkf} &
\hat{\mathbf{p}}^c_{mkf} = \mathbf{S}^{-1} \mathbf{p}^i_{mkf}
\label{eq:transformedSim3}
\end{eqnarray} 
Subsequently the points are transformed to the camera frames and projected to the image plane.
Now, the reprojection error can be computed and used to determine the number of inliers.
If the transformation yields enough inliers, a guided matching is instantiated to search for more correspondences, also between cameras.
Then, $\mathbf{S}$ is optimized by minimizing the reprojection errors of both transformed point sets (\SeeEq{eq:transformedSim3}) in both MKFs.
The optimization is again carried out using g2o and outliers are down-weighted using a Huber kernel.
If more than 20 inliers are retained after optimization, $\mathbf{S}$ is accepted and the loop correction is started.

\paragraph{Loop correction and fusion}
The first step to loop correction is, to correct all MKFs that are connected to the current MKF$_i$, as well as all map points that are part of the local map.
After this step, the local maps spanned by MKF$_i$ and MKF$_c$ should align.
The corrected pose $\hat{\mathbf{M}}_t$ of a MKF is computed by first estimating the relative orientation between pose $\mathbf{M}^{i}_t$ of the current MKF$_i$ and the MKF pose $\mathbf{M}_t$ and subsequent correction using the similarity:
\begin{equation}
\hat{\mathbf{M}}_t = (\mathbf{M}_t \mathbf{M}_t^{i^{-1}}) \mathbf{S}
\label{eq:correctedPose}
\end{equation}
Subsequently, the map points need to be corrected as well.
First, each map point is rotated to the MKF frame using the uncorrected pose.
Then, the point is transformed to the corrected map point position $\hat{\mathbf{p}}$ by applying the inverse of the corrected MKF pose, i.e. the map point is directly transformed back into world coordinates:
\begin{equation}
\hat{\mathbf{p}}
= 
\hat{\mathbf{M}}^{-1}_t (\mathbf{M}_t \mathbf{p})
\label{eq:correctedMapPoint}
\end{equation}
The correction of the local map will result in many point duplications and redundant MKFs.
Thus the same map point fusion procedure presented in \SeeSec{para:mappointfusion} is carried out and the co-visibility graph is updates.

Finally, the essential graph is optimized.
First, all MKF poses $\mathbf{M}_t$ are converted to similarities $\mathbf{S}_t$ by initially setting the scale to 1.0.
Then, the relative pose constrains between all MKFs in the map are computed:
\begin{equation}
\Delta\mathbf{M}_{ij} = \mathbf{M}_{j}\mathbf{M}^{-1}_{i}
\end{equation}
between some MKF $i$ and $j$.
Again, all relative pose constrains are converted to similarities $\Delta\mathbf{S}_{ij}$.
The optimization is carried out over pose-pose constrains and follows the work of \cite{strasdat2010scale}.
The residual that we try to minimize is defined as:
\begin{equation}
\mathbf{r}_{i,j} = \log_{Sim(3)}(\Delta\mathbf{S}_{ij} \mathbf{S}_i \mathbf{S}_j^{-1}) 
\label{eq:residualSimOptim}
\end{equation}
where the $\log$ is the inverse relation of the exponential map (see \cite{strasdat2012local}).
The goal of the optimization is to adjust $\mathbf{S}_i$ and $\mathbf{S}_j$ such that the transformation sequence (back and forth between both MKFs) is as close to the identity as possible.
In the beginning, all residual transformations \SeeEq{eq:residualSimOptim} will be the identity except for the part of the map, that was corrected above.
Then the error is propagated back over all pose-pose constrains during optimization as the similarities $\mathbf{S}_i$ and $\mathbf{S}_j$ are gradually changed to optimally fit the loop closure constrain.

We optimize only the transformations between MKFs that are connected by at least 100 points, i.e. that the edge weight in the co-visibility graph is above 100.
After optimization, all map points are corrected:
\begin{equation}
\hat{\mathbf{p}}
=
\hat{\mathbf{S}}^{-1}_t(\mathbf{M}_t \mathbf{p})
\end{equation}
using the corrected MKF pose $\hat{\mathbf{S}}_t$.
and finally all $\hat{\mathbf{S}}_t$ are converted back to rigid body transformations:
\begin{equation}
\mathbf{M}_t 
=
\begin{bmatrix}
\mathbf{R} & \mathbf{t}/s \\
\mathbf{0}^T & 0
\end{bmatrix}
\end{equation}
where $s$ is the scale.

\begin{figure*}
	\centering
	\subcaptionbox{Trajectory 1\label{fig:loop1Closing}}{
	\includegraphics[clip=true,trim=1mm 1mm 1mm 1mm, width=\textwidth]{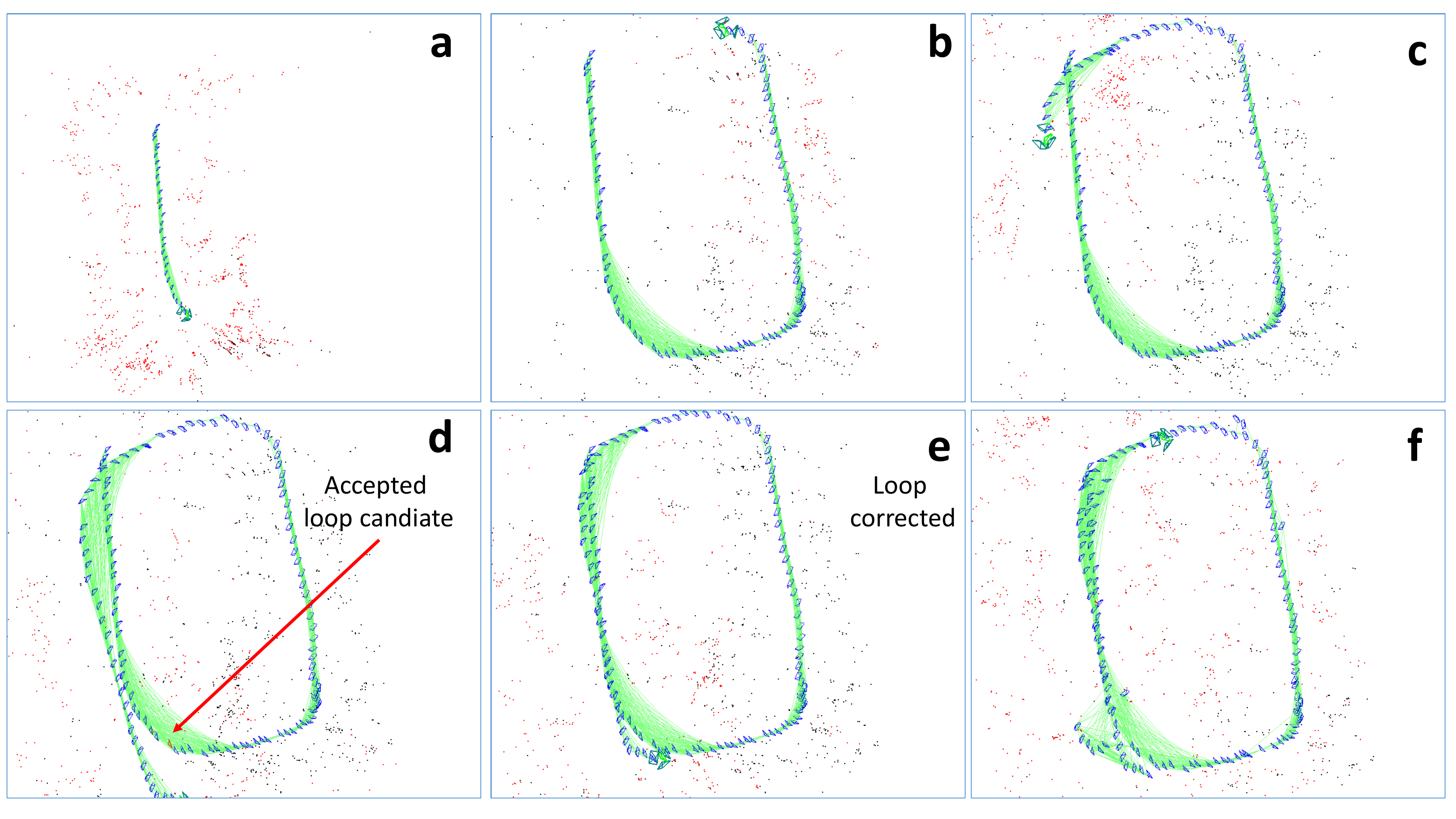}}
	\subcaptionbox{Trajectory 2\label{fig:loop2Closing}}{
	\includegraphics[clip=true,trim=1mm 1mm 1mm 1mm, width=\textwidth]{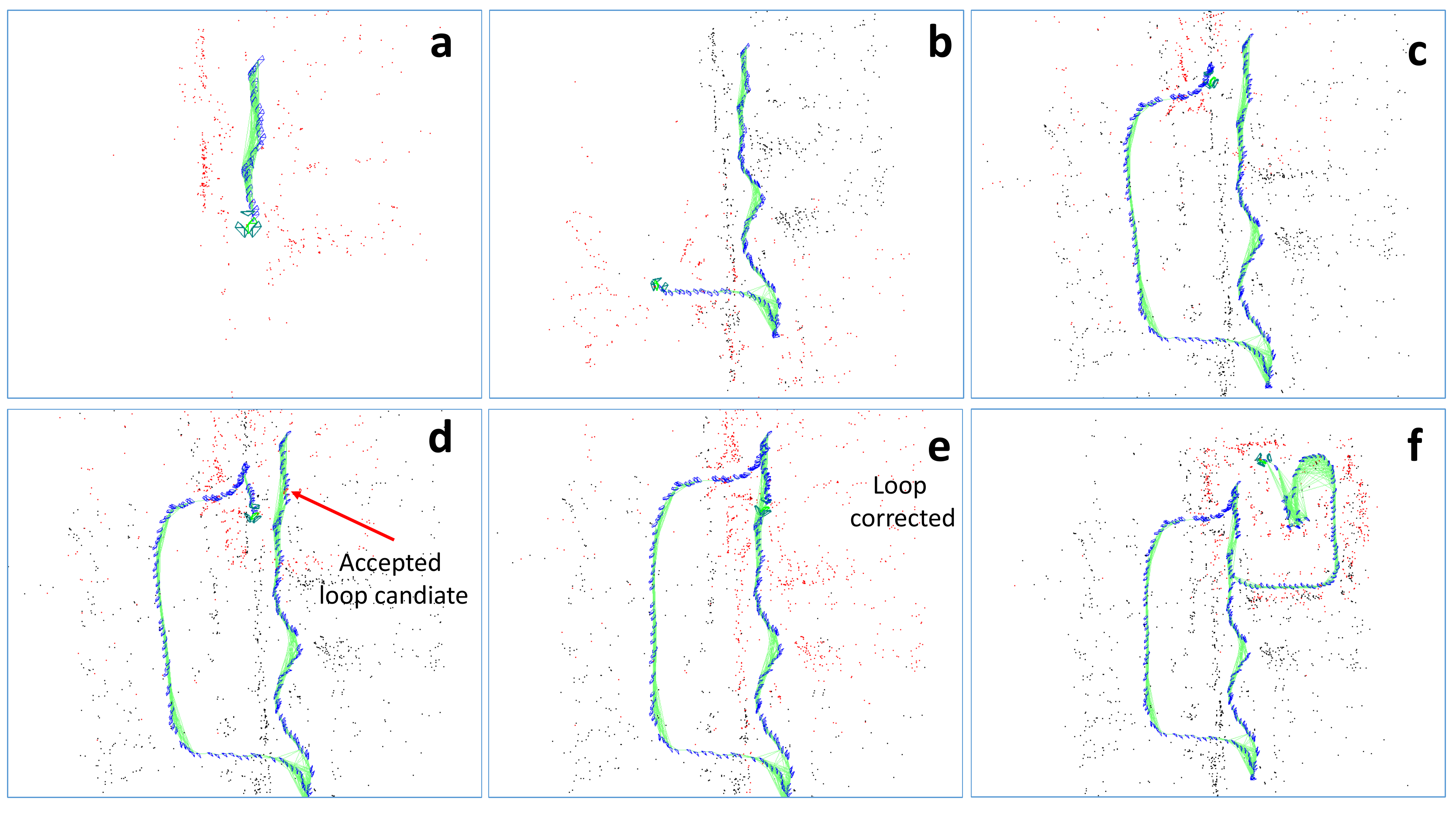}}
	\label{fig:loopClosing}
	\caption{Depicted are two successful loop closures. \textbf{a}-\textbf{b} trajectory and map is build. The green lines represent the co-visibility graph edges with a weight higher than 100. Blue pyramids depict MKF. Red map points is the active map, black points are all map points. \textbf{c} Visually the trajectory crossed itself. However no loop is detected. Although candidates exist, RANSAC does not find a good solution, because the reconstructed geometry is still too different. \textbf{d} a loop candidate is accepted. \textbf{e} The corrected loop after the optimization of the essential graph. \textbf{f} tracking continues using the active map. New MKFs are only created if the baseline is large enough or to few points are tracked.}
\end{figure*}

\section{Experiments and Results}
\label{ch:tracking:sec:results}
To test the performance of the presented MCS SLAM, various tests are performed.
More information about the dataset can be found on \cite{lafida}.
First, we evaluate the impact of using multiple fisheye cameras instead of one, in terms of accuracy, runtime, successfully tracked frames and loop closing.

To evaluate the accuracy of SLAM systems, two metrics are commonly used that compare the estimated camera poses $\mathbf{M}_t$ to a ground truth pose $\mathbf{M}^{gt}_t$ at some time $t$ or an interval $\Delta t$.
The difference between these two poses at time $t$ is given by the relative orientation between them:
\begin{equation}
\mathbf{M}^{rel}_t = {\mathbf{M}^{gt}_t}^{-1} \mathbf{M}_t
\end{equation}
The first metric is called ATE and estimates the root mean squared translation differences between both trajectories.
In order to calculate the absolute error, the two trajectories need to be aligned in advance using a similarity transformation $\mathbf{S}$.
For $N$ pose pairs, the ATE can then be calculated as:
\begin{eqnarray}
\textrm{ATE} 
= &
\sqrt{
\frac{1}{N} \sum_{t=1}^{N} \| \textrm{trans}(\mathbf{M}^{rel}_{t}) \|^2 } \\
= &
\sqrt{
\frac{1}{N} \sum_{t=1}^{N} \| \textrm{trans}({\mathbf{M}^{gt}_{t}}^{-1} \mathbf{S} \mathbf{M}_t )\|^2}
\label{eq:ATE}
\end{eqnarray} 
where "$\textrm{trans}$" returns the translational component of the transformation matrix $\mathbf{M}$.

The second metric is called RPE and allows to evaluate the local accuracy and drift of the trajectory over some time interval $\Delta$.
Thus we can calculate $M=N-\Delta$ relative orientation errors along the trajectory.
The RPE at time step $t$ can be defined by:
\begin{eqnarray}
\textrm{RPE}(\Delta) 
= &
\sqrt{\frac{1}{M} \sum_{t=1}^{M} \| \textrm{trans}(\mathbf{M}^{rel}_t) \|^2 }
\label{eq:RPE}
\end{eqnarray}
but this time the relative transformation is defined as:
\begin{equation}
\mathbf{M}^{rel}_t
=
({\mathbf{M}^{gt}_t}^{-1}\mathbf{M}^{gt}_{t+\Delta})^{-1} (\mathbf{M}^{-1}_t \mathbf{M}_{t+\Delta})
\end{equation}
To calculate the relative error of subsequent poses we set $\Delta = 1$.
In the case of ATE only the translation is evaluated.
For the relative error, we can also evaluate the rotational accuracy.
This is done by replacing the "$\textrm{trans}$" with a function that returns the Rodriguez vector of the rotation matrix in $\mathbf{M}$.
%The drift is evaluated per second, i.e. $\Delta=25$ if the cameras run at 25Hz.

Each trajectory is evaluated five times, i.e. the SLAM algorithms are used five times to estimate the camera trajectory.
All accuracies and run-times are calculated as the median value over the five runs.

\subsection{Single- vs. Multi-Camera SLAM}
\label{ch:tracking:sec:singlevsmulti}
First, we align the KFs or MKFs respectively, by estimating a similarity transformation between ground truth and SLAM trajectory.
Then, the ATE (\SeeEq{eq:ATE}) is evaluated for all trajectories.
The results are depicted in \SeeTable{tab:ATESLAM}.
Obviously, MultiCol-SLAM significantly outperforms its single camera pendant in terms of Keyframe accuracy.
One explanation of the large performance gap is the simple initialization of the single camera SLAM.
The authors of ORB-SLAM proposed an initialization based on homography and fundamental matrix estimation.
Both matrices can not be readily computed for the camera model employed in this work.
Thus we simply initialize the single camera SLAM by estimating the essential matrix and selecting a solution based on a threshold on the magnitude of the translation vector, which is obviously less robust than the method proposed in \cite{mur2014orb}.

To get a measure of the local accuracy, we also estimate the RPE (\SeeEq{eq:RPE}) for all trajectories and all poses by setting $\Delta = 1$.
The trajectories do not need to be aligned in this case.
The accuracies for translation and rotation are depicted in \SeeTable{tab:RPEDELTA1}.
Still, using multi-cameras yields a better performance, especially for the translation. 
The rotational components show a similar trend, but the differences are less prominent.
The rotational accuracy for the two lasertracker trajectories is a lot better because the trajectory has only little rotation of the camera system about the up-axis.
On average the rotational accuracy of the MCS is about 0.5-1.5$^\circ$ and the translational component, depending on the walking speed and scene between 1.0-2.5 cm.

\begin{figure*}
	\resizebox{\textwidth}{!}{%
\subcaptionbox{\textbf{ATE}\label{tab:ATESLAM}}{	
\centering
\begin{tabular}{l|c|c|}
	\cline{2-3}
	& Single fisheye camera & Multi-fisheye camera system \\ \cline{2-3} 
	& {[}cm{]}              & {[}cm{]}                    \\ \hline
	\multicolumn{1}{|l|}{Laser 1}             & 31.0                  & 1.4                         \\ \hline
	\multicolumn{1}{|l|}{Laser 2 fast}        & 28.1                  & 5.3                         \\ \hline
	\multicolumn{1}{|l|}{Indoor 1 stat. env.} & 32.4                  & 2.1                         \\ \hline
	\multicolumn{1}{|l|}{Indoor 2 dyn. env.}  & 13.3                  & 1.8                         \\ \hline
	\multicolumn{1}{|l|}{Outdoor 1 dyn. env}  & (X)                   & 3.6                         \\ \hline
\end{tabular}
}
\subcaptionbox{\textbf{RPE}\label{tab:RPEDELTA1}}{	
	\centering	
	\begin{tabular}{|c|c|}
		\hline
		Single fisheye camera & Multi-fisheye camera system \\ \hline
		{[}cm{]}/{[}deg{]}  & {[}cm{]}/{[}deg{]}   \\ \hline
		1.95/0.32 & 1.2/0.33                         \\ \hline
		2.6/0.31  & 2.7/0.56                      \\ \hline
		2.8/1.72 & 1.1/1.54                      \\ \hline
		2.8/2.04 & 1.1/1.78                        \\ \hline
		(X) & 2.3/1.28                        \\ \hline
	\end{tabular}
}}
	\caption{(a) Median KF and MKF \textbf{ATE}s for single and multi-camera SLAM respectively. The translational accuracy was calculated after 7DoF alignment between ground truth and estimated frames. (b) \textbf{RPE} for single and multi-fisheye camera SLAM. Here we set $\Delta=1$, i.e. the frame two frame tracking accuracy is estimated.
	(X) means that tracking failed at some point and a significant part of the trajectory was not tracked.}
\end{figure*}

\begin{table*}[]
	\centering
	%\resizebox{\textwidth}{!}{%
\begin{tabular}{|l|l|c|c|}
	\hline
	Thread                                         & Operation          & Single fisheye camera & Multi-fisheye camera system \\ \hline
	\multirow{4}{*}{Tracking}                      & Frame creation     & 12/12/2               & 21/23/5                     \\ \cline{2-4} 
	& Pose estimation    & 4/4/1                 & 3/4/3                       \\ \cline{2-4} 
	& Track local map    & 5/5/2                 & 4/5/4                       \\ \cline{2-4} 
	& Total              & \textbf{21}           & \textbf{28}                 \\ \hline
	\multicolumn{1}{|c|}{\multirow{4}{*}{Mapping}} & Map point creation & 27/26/10              & 96/96/17                    \\ \cline{2-4} 
	\multicolumn{1}{|c|}{}                         & Map point fusion   & 11/11/6               & 11/12/9                     \\ \cline{2-4} 
	\multicolumn{1}{|c|}{}                         & Local BA           & 185/198/125           & 170/174/65                  \\ \cline{2-4} 
	\multicolumn{1}{|c|}{}                         & Total              & \textbf{223}          & \textbf{277}                \\ \hline
\end{tabular}
%}
	\caption{Depicted is the time (mean, median and standard deviation) for each step in different threads. In case of Single camera SLAM, we extracted 1000 features. In case of multi-camera SLAM 400 features per camera are extracted and the extraction is performed in parallel. The pose estimation is slower in the single camera case, as no analytical Jacobian was provided.}
	\label{tab:timingsSLAM}
\end{table*}

\section{Summary And Conclusion}
\label{ch:tracking:sec:conclusion}
In this paper MultiCol-SLAM, a real-time multi-fisheye camera SLAM system was proposed.
First, we recapitulated the current state-of-the-art in the field and argued why keyframe-based approaches outperform filter-based SLAM systems.
Then we subsumed the MultiCol model and detailed our contributions.
Subsequently, we elaborately detailed our framework that builds upon ORB-SLAM and is divided into several threads running in parallel.
Finally, all proposed modules were examined using accurate ground-truth data and it showed, that using multi-camera systems helps to improve the accuracy and robustness of SLAM in challenging environments.

In addition, we make the proposed SLAM system available to the public (\url{https://github.com/urbste/MultiCol-SLAM}) and hope that it helps to further encourage research in multi-camera egomotion estimation and related topics.
%% ==============================

\section*{Acknowledgment}
This project was partially funded by the DFG research group FG 1546 "Computer-Aided Collaborative Subway Track Planning in Multi-Scale 3D City and Building Models".
\begin{spacing}{0.8}% tune the size by altering the parameter
	{\small \bibliography{MultiColSLAM_bib}} % Include your own bibliography (*.bib), style is given in isprs.cls
\end{spacing}

\end{document}